\pdfoutput=1

\documentclass[11pt]{article}

\usepackage[]{ACL2023}

\usepackage{times}
\usepackage{latexsym}
\usepackage{enumitem}
\usepackage{url, hyperref}
\usepackage[T1]{fontenc}

\usepackage[utf8]{inputenc}

\usepackage{microtype}

\usepackage{inconsolata}

\usepackage{danudefs}
\usepackage{booktabs}
\usepackage{graphicx}
\usepackage{amsmath, amsfonts, amssymb, amsthm, mathtools}
\usepackage{algorithm, algorithmic}
\title{Improving Pre-trained Language Model Sensitivity via Mask Specific losses: \\
A case study on Biomedical NER}

 \author{Micheal Abaho$^{1,2}$ \\
  \\\And \kern+2.5em
  Danushka Bollegala$^{1}$ \\
  \\\And \kern+4.5em
  Gary Leeming$^{1,2}$ \\ 
  \kern1em $^1$University of Liverpool, United Kingdom \\
     \kern1em $^{2}$Civic Health Innovation Labs \\
     \kern+1em \texttt{\{micheal.abaho,danushka,gary.leeming,d.joyce,buchan\}@liverpool.ac.uk} 
  \\\And \kern+2.5em
  Dan Joyce$^{1,2}$ \\
  \\\And
  Iain E Buchan$^{1,2}$ \\
\\}

\begin{document}
\maketitle
\begin{abstract}
Adapting language models (LMs) to novel domains 
is often achieved through fine-tuning a pre-trained LM (PLM) on domain-specific data. Fine-tuning introduces new knowledge into an LM, enabling it to comprehend and efficiently perform a target domain task. Fine-tuning can however be inadvertently insensitive if it ignores the wide array of disparities (e.g in word meaning) between source and target domains. For instance, words such as \textit{chronic} and \textit{pressure} may be treated lightly in social conversations, however, clinically, these words are usually an expression of concern. To address insensitive fine-tuning, we propose  Mask Specific Language Modeling (MSLM), an approach that efficiently acquires target domain knowledge by appropriately weighting the importance of domain-specific terms (DS-terms) during fine-tuning. MSLM jointly masks DS-terms and generic words, then learns mask-specific losses by ensuring LMs incur larger penalties for inaccurately predicting DS-terms compared to generic words. 
Results of our analysis show that MSLM improves LMs sensitivity and detection of DS-terms. We empirically show that an optimal masking rate not only depends on the LM, but also on the dataset and the length of sequences. Our proposed masking strategy outperforms advanced masking strategies such as span- and PMI-based masking. 
\end{abstract}

\section{Introduction}

Fine-tuning is the prevailing practice for adapting an LM to a new domain. A plethora of research works ranging from task-generalization \cite{claudino2018crossfit, peters2019tune, peng2019transfer}, to few-shot learning \cite{gao2020making, mccann2018natural} to in-context tuning \cite{chen2021meta} all unanimously credit fine-tuning for the state-of-the-art results across a diverse set of NLP tasks. Despite its remarkable strides, fine-tuning has been reasonably criticised for its instability and brittleness by a few pockets of NLP researchers \cite{mosbach2020stability, lee2019mixout, dodge2020fine}. \citet{lee2019mixout, dodge2020fine} attributed fine-tuning's instability to catastrophic forgetting and small sized datasets, and most recently \citet{mosbach2020stability} exposed the optimization challenges encountered during fine-tuning LMs. 

\begin{table}[t]
\resizebox{\columnwidth}{!}{
\begin{tabular}{@{}ll@{}}
\toprule
\multicolumn{2}{c}{\textbf{Social Conversation}}                                                                                                                                            \\ \midrule
Dan:  & Hi Gary, how was your week?                                                                                                 \vspace{0.5em}                                                         \\
Gary: & \begin{tabular}[c]{@{}l@{}}It has ended well but I had a lot of \colorbox{bubblegum}{\textbf{pressure}} \\ throughout the week to meet a deadline. I \\ felt like I would get \colorbox{bubblegum}{\textbf{attacked}} by colleagues.\end{tabular} \\ \midrule
\multicolumn{2}{c}{\textbf{Clinical Conversation}}                                                                                                                                          \\ \midrule
Dan:  & Hi Gary, how was your week?                                                                    \vspace{0.5em}                                                                                      \\
Gary: & \begin{tabular}[c]{@{}l@{}}It has ended well but my \colorbox{cadmiumred}{\textbf{pressure}} was high \\ throughout the week. I felt like I would get \\ an \colorbox{cadmiumred}{\textbf{attack}}.\end{tabular}                                     \\ \bottomrule \bottomrule
\end{tabular}
}
\caption{Comparing the sensitivity of two words in two different conversations (Social and Clinical setting). The brighter the colored boxes wrapping the words, the more concerning for the respective conversation.}
\label{tab:comp_sens}
\vspace{-1em}
\end{table}

\begin{figure*}[t]
\centering
\includegraphics[width=0.95\textwidth]{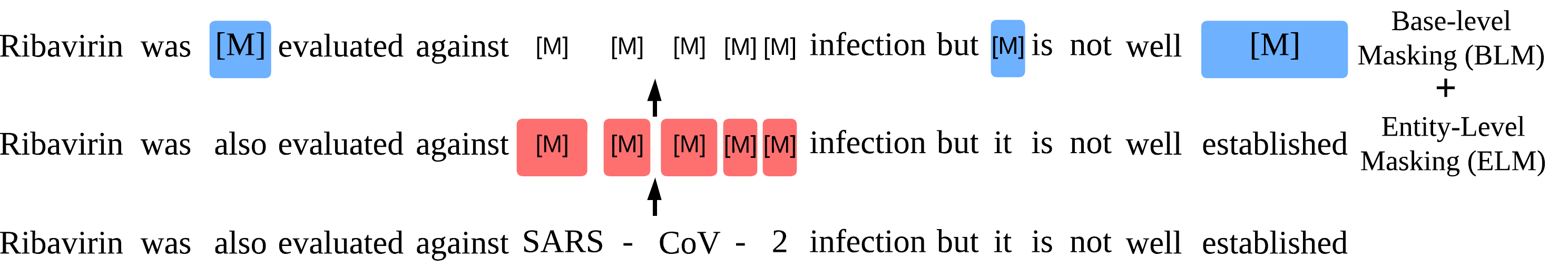}
\caption{Joint ELM-BLM masking of tokens in an input sequence.}
\label{fig:mask_ing}
\vspace{-1em}
\end{figure*}

It is notable that, across all prior critics, the focus and attention has been strongly directed towards the performance of these LMs, and very limited attention has been paid towards the sensitivity and domain-specific knowledge these LMs pickup during fine-tuning. 
There is usually a wide range of disparities between the source (used for pre-training) and the target (used for fine-tuning) domains. 
Some of these may include but not limited to, word meaning \cite{navigli2009word, zhou2021learning}, word intensity (strength or potency of a word in given domain) \cite{yin2020construction, baek2022prosodic} and abbreviation disambiguation \cite{wu2015clinical}. If these disparities are not properly catered for, fine-tuning can easily become an underwhelming adaptation process 
and insensitive to specialised target domains. For instance, words such as \textit{chronic}, \textit{pressure} and \textit{attack} will often be treated lightly in social conversations, however, clinically these words are usually a cause of concern. For example, we notice that, whereas the words ``\textit{pressure}'' and ``\textit{attack}'' are mentioned by the respondent in both the social and clinical contexts in \autoref{tab:comp_sens}, they definitely require more attention in the clinical conversation, hence the questioner ``Dan'' ought to be more sensitive to the respondent ``Gary''. 


In this work, we address the domain sensitive fine-tuning (DSFT) discussed in the previous paragraph. We use insensitivity in our context to imply the below par awareness of DS-terms, rather than language-insensitivity as it pertains to human feelings. 
We investigate the hypothesis: \textbf{``The awareness of or sensitivity of PLMs 
towards DS-terms can be appropriately elevated without hurting their downstream performance''}.

In order to strategically increase an LM's awareness of DS-terms, we revisit the language understanding and generation phenomenon of Mask Language Modeling (MLM) \cite{devlin2018bert}. We modify MLMs to up-weight the significance of masked  DS-term tokens such that the attention towards them is relatively larger than that towards masked non DS-term tokens. In doing so, we introduce the notion of \textbf{``mask-specific loss''}, which we compute using appropriately assigned weights that are computed using a strategy similar to the one \citealp{mosbach2020stability} used to address class imbalance. We further introduce entity recognition and entity classification objectives to collectively contribute towards a cross entropy loss with an aim to enhance the ability of a model to detect mentions. We refer to this approach as Mask-Specific Language Modeling (MSLM\footnote{\url{https://github.com/mykelismyname/MSLM}}).


Using the biomedical domain as our test bed, we evaluate how well MSLM can perform when tasked to extract clinical entities from a host of datasets within the Biomedical Language Understanding \& Reasoning Benchmark (BLURB) \cite{gu2021domain}.
To study the effectiveness of MSLM, we do not simply compare the perplexity of our sensitive models to the vanilla models, instead, we proceed to check confidence scores with which the two sets of models predict DS-terms. 
We assess the impact of our proposed masking strategy by varying the masking rate and lengths of input sequences and monitoring their influence on the LMs prediction results. In addition,
we study how this masking strategy compares to other advanced strategies such as PMI (Pointwise Mutual Information) \cite{levine2020pmi} and Span \cite{joshi2020spanbert}.
Our experiments demonstrate (a) a performance improvement in extraction of exact mentions of named entities, (b) the influence the masking rate and sequence lengths has on prediction performance, and  (c) the superiority of the proposed masking strategy over other advanced masking strategies.

\section{Mask-Specific Language Modeling}
In designing our approach, we draw lessons from two prior tested and proven phenomena:
(1) MLMs are effective in learning representations for sub-tokens, words \cite{devlin2018bert}, phrases \cite{sun2019ernie} and spans \cite{levine2020pmi, joshi2020spanbert}; and 
(2) high prediction rates (proportion of tokens to be predicted) substantively affect optimization, i.e. they increase training signals, which subsequently boost performance \cite{wettig2022should}.
We refer to these two phenomena respectively as the \emph{MLM-effect} and the \emph{High-prediction-effect} in the remainder of this paper.

\subsection{Masking}
\label{sec:masking}

Randomly replacing a proportion of tokens in a sentence with $\mathrm{[MASK]}$ tokens (Base level Masking ~\cite[\textbf{BLM};][]{devlin2018bert}) intuitively enables LMs to learn the bi-directional context that often surrounds words in written language text.

Because certain spans of words are best understood when all of their constituted words are written together to denote a named entity such as a person, an organisation and a location, replacing named entity spans with $\mathrm{[MASK]}$ tokens (Entity level Masking ~\cite[\textbf{ELM};][]{sun2019ernie, abaho2022position}) has also proven to be effective in learning contextualised representations for these entities.

We leverage benefits of the two above strategies and propose a new masking strategy, ``Joint ELM-BLM'' shown in \autoref{fig:mask_ing}. On its own, ELM would help enrich an LM with contextual knowledge necessary in discriminating our targeted DS-terms, however, when exploiting the \emph{MLM-effect}, we consider it necessary to avoid tightly coupling the LM's weights onto these DS-terms. We therefore utilise BLM to preserve a PLM's inherent domain and generic knowledge. More so, we avoid the assumption that 15\% masking rate is optimal \cite{devlin2018bert} and instead explore a spectrum of rates to find an optimum. In our experimental setup, we ensure that BLM- and ELM-masked sets are disjoint sets of tokens.
 




Besides datasets with annotations of DS-terms (clinical entities), we assume access to a Biomedical PLM denoted as $\mathrm{Enc_{PLM}}$. This LM can be used for encoding each input sequence $s$ of $n$ tokens to obtain $\mat{H}$, a matrix of $n$ vectors as shown in \eqref{eq:enc}.

\begin{align}
    \mat{H} = \mathrm{Enc_{PLM}}(x_1, \ldots, \mathrm{[MASK]}_i, \ldots, x_n)
    \label{eq:enc}
\end{align}

\subsubsection{Mask specific losses}
\label{sec:msl}
The main goal in our approach is to strategically increase a PLM's sensitivity towards DS-terms while simultaneously retaining sufficient knowledge of generic terms.
The first attempt in achieving this is masking DS-terms along with generic terms as discussed in \textsection\ref{sec:masking}. 

To further achieve our goal, we introduce the idea of mask specific losses, which essentially aims to impose larger penalties on the model for inaccuracies in predicting corrupted (masked) DS-terms compared to the corrupted generic terms. 

Typically, instance-specific losses are computed by re-scaling weights for each possible class in the label space \cite{wang2017learning, cui2019class}, however, in this case, rather than classes, we have ELM- and BLM-masked tokens as well as un-masked tokens. 
To compute the weights assigned to the tokens in our masked input, we firstly obtain the number of ELM- and BLM-masked tokens within the training dataset and denote them as $\mathrm{N_{ELM}}$ and $\mathrm{N_{BLM}}$ respectively.
A mask specific weight is computed for each of the mask types (ELM \& BLM), as the difference between 1 and the the corresponding mask type probability (i.e. the mask type's distribution out of the total mask types distribution), given by \eqref{eq:distribution_weight}. 
The final mask specific weight is obtained as the softmax over the mask specific weights from previous step as given by \eqref{eq:sensitivity_threshold}.

\begin{table*}[!t]
\centering
\resizebox{\textwidth}{!}{
\begin{tabular}{@{}lcccccc@{}}
\toprule
 & \begin{tabular}[c]{@{}c@{}}\#Sents\\ {\small Train | Val | Test}\end{tabular} & \#Classes & AvgSentLen & \multicolumn{1}{c}{\begin{tabular}[c]{@{}c@{}}\#Ments\\ {\small Train | Val | Test}\end{tabular}} & AvgMents & AvgMentsLen \\ \midrule
BC2GM &        12632 | 2531 | 5065    & 2          &  25.17          & 15197 | 3061 | 632  & 1.20 & 2.4 \\
NCBI-disease &  5432 | 923 | 942      & 2          &  25.24          & 5134 | 787 | 960    & 0.95 & 2.2 \\  
BC5CDR-chem &  4812 | 4602 | 4582     & 2          &  25.75          & 5385 | 5203 | 5347  & 1.12 & 1.3\\ 
\bottomrule
\end{tabular}
}
\caption{Dataset statistics. \#Sents is the number of sentences and \#Ments is the number of DS-term mentions, AvgSentLen is the average length of sentences, AvgMents is the average number of DS-terms mentioned per sentence obtained as (\# of train Ent\_Ments)/(\# of train sents). Full table with all datasets in \ref{tab:full_dataset_stats} in the Appendix.}
\label{tab:dataset_stats}
\vspace{-1em}
\end{table*}


\begin{align}
        w_x = 1-\frac{\mathrm{N_{x}}}{\sum_{x \in \{\mathrm{BLM,ELM}\}}\mathrm{N_x}} 
        \label{eq:distribution_weight}
\end{align}

\begin{align}
\label{eq:penalty_constraint}
\begin{split}
    w_{\mathrm{BLM}} &=
\left\{
	\begin{array}{ll}
		0.5  & \mbox{if } w_{\mathrm{BLM}} > 0.5 \\
		w_{\mathrm{BLM}}
	\end{array}
\right. \\
  w_{\mathrm{ELM}} &=
\left\{
	\begin{array}{ll}
		0.5  & \mbox{if } w_{\mathrm{ELM}} < 0.5 \\
		w_{\mathrm{ELM}}
	\end{array}
\right. 
\end{split} \\
 w &= ([w_{\mathrm{BLM}}, w_{\mathrm{ELM}}])
\\
 \label{eq:sensitivity_threshold}
   \vec{w} &= \mathrm{softmax}(w)
\end{align}

In order to elevate the sensitivity towards DS-terms but equally avoid overfitting onto them, we introduce a sensitivity threshold, which is used to encourage the ELM-masked tokens related weight ($w_{\mathrm{ELM}}$) and also to carefully suppress the BLM-masked tokens related weight ($w_{\mathrm{BLM}}$). 
Because of the sporadic nature of the mentions of DS-terms within the dataset, the distribution of ELM-masked tokens will typically be lower than that of BLM-masked tokens, in other words not every input sequence will have a mention of DS-term/s, while every input sequence will have tokens that are subject to BLM. We therefore set the sensitivity threshold to 0.5 to force a balance in their probability distribution (i.e. implying that BLM and ELM are equally likely to occur for a given input sequence). We then ensure that $w_{\mathrm{BLM}}$ never rises above this threshold and similarly, $w_{\mathrm{ELM}}$ should never fall below that threshold as shown in \eqref{eq:sensitivity_threshold}.

The normalized weight vector $\vec{w}$ is used to compute the MSLM loss ($L_{\mathrm{MSLM}}$) during the prediction of the masked tokens ${x_i}$ as given by \eqref{eq:mslm_loss}.
\begin{align}
    \label{eq:mslm_loss}
    L_{\mathrm{MSLM}} = -\sum w_i^{(x)} \log P(x_i|s)
\end{align}
Here, $w_i^{(x)} \in \vec{w}$ is a mask-specific weight for a masked token $x_i$ that lies within the sequence $s$.

\subsection{Entity detection and Classification}
Because the biomedical domain has many classification schemes that are used in categorizing clinical entities \cite{jackson2018cogstack, gu2021domain}, we maximize the \emph{High-prediction-effect} by formulating an entity recognition and classification task. The idea behind this is, the more predictions a model has to make (both in predicting masked-out tokens as well as classifying unmasked entities), the more signals it would get through computing gradients during optimization.
The entity recognition task is defined below.

\paragraph{Task formulation:} Given a sentence $s = \{x_i\}_{i=1}^n$ of $n$ tokens, where each $x_i$ is tagged with a BIO label \cite{sang1999representing}, we build a model that can accurately extract entities $\{e_i^{(s)}\}_{i=1}^N$ mentioned in $s$. We obtain a probability distribution across all BIO labels as given by \eqref{eq:entity_detection}.
\begin{align}
    \label{eq:entity_detection}
    \hat{y}_i = \mathrm{softmax}(f(\vec{h_i} \circ \vec{W}^{(ed)}))
\end{align}

Here, $f$ is a nonlinear function, $\circ$ denotes the vector concatenation and $\vec{W}^{(ed)} \in \R^{1 \times k}$ is a trainable weight vector, $h_i \in \mat{H}$. In addition to $L_{\mathrm{MSLM}}$, we compute an entity detection loss given by \eqref{eq:entity_detection_loss}.
\begin{align}
  \label{eq:entity_detection_loss}
    L_{\mathrm{ED}} = -\sum_{i=1}^{n}\sum_{j \in \mathrm{BIO}} y_{i,j} \log \hat{y}_{i,j}
\end{align}

\paragraph{Entity Linking/Classification loss:}
Given a detected entity, we obtain an entity span representation in \eqref{eq:entity_rep}, and compute probability distribution across all entity types $E$ in \eqref{eq:entity_classification},

\begin{figure*}[!t]
\centering
\includegraphics[width=0.95\textwidth]{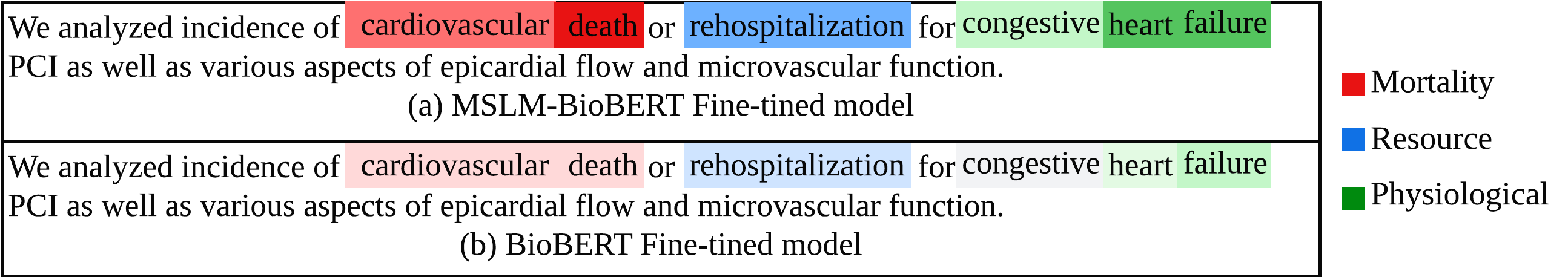}
\caption{Visualization of the confidence score with which different DS-terms belonging to different outcome (within the EBM-NLP dataset) are predicted. The color intensity increases with the confidence score.}
\label{fig:confidence_scores}
\end{figure*}

\begin{align}
\label{eq:entity_rep}
 \vec{e_m} = \mathrm{meanpool}(h_i,\ldots,h_M) 
\end{align}

where the entity $m$ has $1$ to $M$ tokens. 

\begin{align}
    \label{eq:entity_classification}
    \hat{y}_m^l = \mathrm{softmax}(f(\vec{e_m} \circ \vec{W}^{(ec)}))
\end{align}

where $f$ is a non-linear function and $\vec{W}^{(ec)} \in \R^{1 \times d}$ is a trainable weight vector. We introduced the task specific trainable parameters $\vec{W}^{(ed)}$ and $\vec{W}^{(ec)}$ to enrich the MLM representations that are used in the token class prediction layer and the entity class prediction layer respectively. The MSLM loss would benefit from the extra knowledge brought in from these parameters during optimization. Task type trainable parameters have proven to be beneficial in prior work \cite{yao2019docred, eberts2021end}. 

The classification loss is given by \eqref{eq:class-loss}.
\begin{align}
\label{eq:class-loss}
    L_{\mathrm{EL}} = -\sum_{l \in E} y_m^l \log \hat{y}_m^l
\end{align}

\paragraph{Model loss:} We optimize the joint loss of all three cross-entropy losses as given in \eqref{eq:model_loss}.
\begin{align}
\label{eq:model_loss}
   L = L_{\mathrm{MSLM}} + L_{\mathrm{ED}} + L_{\mathrm{EL}}
\end{align}


\section{Experiments}
To evaluate MSLM, we initialize multiple biomedical LMs which were pre-trained on massive collections of publicly available scientific literature in PubMed. Compared LMs include
\textbf{BioBERT}~\cite{lee2020biobert}, \textbf{SciBERT}~\cite{beltagy2019scibert}, \textbf{PubMedBERT}~\cite{gu2021domain} and  \textbf{BioELECTRA}~\cite{raj2021bioelectra}.  

\paragraph{Datasets:}
To facilitate our investigation, we use Named Entity Recognition (NER) datasets within the BLURB benchmark \cite{gu2021domain}. These include \textbf{NCBI-disease} containing 6892 disease mentions linked to 790 distinct disease concepts, \textbf{BC5CDR-Disease} \& \textbf{BC5CDR-Chemical} containing mentions of diseases and chemicals in 1,500 PubMed articles, \textbf{BC2GM} containing 20,000 sentences with gene mentions, \textbf{JNLPBA} containing 2,000 PubMed abstracts with mentions of molecular biology-related entities such as DNA and \textbf{EBM-NLP} containing 5,000 PubMed clinical trial abstracts with mentions of the PICO elements (We specifically use the version with denoised outcome annotations as used by \newcite{abaho2019correcting, abaho2021detect}).

\paragraph{Metrics:}
We use an exact match (EM) score metric to measure the sensitivity towards DS-terms. EM counts a prediction of an entire entity as 1 if and only if it completely matches the correct answer, both in terms of the precise boundary of the DS-term mention as well as the term's classification. Furthermore, we measure macro-F1 score for NER performance \cite{hajic2009conll} and perplexity of the models to monitor how well the models adapt to and comprehend the domain datasets. 

\begin{table}[!b]
\resizebox{\columnwidth}{!}{
\begin{tabular}{@{}llccc@{}}
\toprule
             &            & \textbf{Vanilla}  & \begin{tabular}[c]{@{}c@{}}\textbf{MSLM}\\ {\small ELM=1,BLM=0.075}\end{tabular} \\ \midrule
BC2GM        & BioBERT    &   88.4           &   \textbf{90.3$_{\pm 0.5}$}                                                                    \\
             & PubMedBERT &  86.8           & \textbf{89.8$_{\pm 0.4}$}                                                                      \\
             & BioELECTRA &  87.6               &   89.1$_{\pm 0.2}$                                                                      \\
             & SciBERT &  85.7              &  \textbf{87.1$_{\pm 0.4}$}   
                                                         \\
NCBI-disease & BioBERT    &    89.1         &     \textbf{90.1$_{\pm 0.1}$}                                                                  \\
             & PubMedBERT &  \textbf{89.9}           &  \textbf{89.9$_{\pm 0.2}$}                                                                      \\
             & BioELECTRA &  88.5               &    \textbf{88.9$_{\pm 0.2}$}                                                                    \\
             & SciBERT &  88.4              & \textbf{89.9$_{\pm 0.1}$}    
                                                         \\

BC5DCR-chem      & BioBERT    &   93.3      &  \textbf{94.0$_{\pm 0.2}$}                                                                      \\
             & PubMedBERT &  94.0           & \textbf{94.4$_{\pm 0.2}$}                                                                        \\
             & BioELECTRA &  90.8              &  \textbf{94.0$_{\pm 0.2}$}                                                                  \\
             & SciBERT &   90.7               & \textbf{93.7$_{\pm 0.2}$}   
                                                         \\

EBM-NLP      & BioBERT    &  64.3        &   \textbf{75.4$_{\pm 0.4}$}                                                                    \\
             & PubMedBERT &  65.5        &   \textbf{76.2$_{\pm 0.3}$}                                                                    \\
             & BioELECTRA &  63.7            &   \textbf{73.2$_{\pm 0.3}$}                                                                       \\
             & SciBERT &  69.7      &   \textbf{73.4 $_{\pm 0.2}$}  
                                                         \\ \bottomrule
\end{tabular}
}
\caption{Exact match (EM) scores obtained when MSLM (ELM=100\%, BLM=7.5\%) is initialized with various biomedical PLMs. Average scores across 5 runs and their standard deviation are reported for the MSLM models which are compared against Vanilla versions of the LMs. Best results are in bold and full results are provided in \autoref{tab:full_exact_match_scores} in Appendix.}
\label{tab:exact_match_scores}
\vspace{-1em}
\end{table}

\paragraph{Setup:}
\label{sec:setup}
Two important factors in our setup include, (1) we establish ELM rate with respect to the total number of DS-terms mentioned in an input sequence rather than all input sequence tokens. For example, if the number of DS-terms in a sequence $s$ is denoted as $\mathrm{DS}_s$ and $\mathrm{DS}_s = 4$, an ELM of 25\% implies $0.25 \times 4 = 1$, hence 1 out of the 4 DS-terms are randomly masked. Whenever this computation returns a decimal value, we round off the value upward to the 
nearest integer (e.g if $\mathrm{DS}_s = 3$, and ELM=25\%, $0.25 \times 3 = 0.75$, which will be rounded off to 1), (2) Since ELM consumes a portion of the masking budget as explained above, we halve the conventional 15\% rate to get a BLM rate of 7.5\%. Furthermore, high masking rates are not favourable for moderately-sized (ca. 125M parameters) LMs \cite{wettig2022should} such as the ones we use in this paper. This constraint is used in our initial set of experiments, however, later on, we explore how varying both BLM and ELM rates would affect model performance especially because the average length of sentences and DS-term mentions varies across different datasets listed in \autoref{tab:dataset_stats}.

\paragraph{Implementation details:}
The infrastructure used in our experiments includes, PyTorch 2.0 for developing MSLM and two GPU machines, a 48G NVIDIA RTX A6000 and a 28G N-series (NC6s\_v3) Azure Virtual Machine. The two GPUs are not used to concurrently run the same experiment but to run different experiments in parallel. 
Results reported are based on testing performance. Dataset statistics are included in \autoref{tab:dataset_stats}.

\begin{figure*}[!t]
\centering
\includegraphics[width=\textwidth, height=3.7cm]{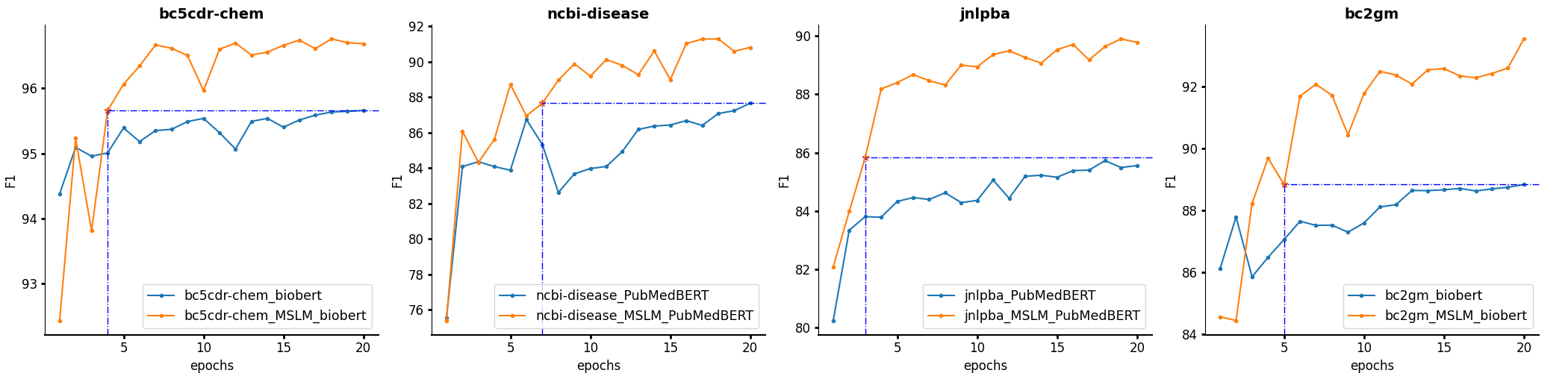}
\caption{Downstream NER F1 performance of the vanilla and the MSLM-fine-tuned models (i.e. DSFT models). ELM and BLM rates used in \textsection\ref{sec:sensitivity_dst} are maintained. More plots in Appendix \ref{sec:dsft_destructive}.}
\label{fig:dsft_f1_scores}
\end{figure*}

\subsection{Sensitivity towards DS-terms}
\label{sec:sensitivity_dst}

To investigate the sensitivity of MSLM-fine-tuned models, we evaluate two metrics: (a) the confidence in the models predictions, and (b) the EM score of the predictions. 
With the former, we visualize the softmax probabilities (which we also refer to as confidence scores) with which model predicts DS-terms using the heatmap in \autoref{fig:confidence_scores}. For demonstration purposes, we use the EBM-NLP dataset since it has multiple classes in comparison to the other datasets. As observed in \autoref{fig:confidence_scores}, despite both sets of models predicting the correct classes for the 3 DS-terms, cardiovascular death (Mortality outcome), rehospitalization (Resource-use outcome) and congestive heart failure (Physiological outcome), the confidence score with which the model predicts classes for the DS-terms is visibly higher for MSLM-BioBERT models. 

\autoref{tab:exact_match_scores} reports EM scores, which are indicative of the model performance in detecting full or exact mentions of DS-terms. We notice that, MSLM improves the performance (+3.2 percentage points on average) with which LMs detect full mentions. Most notably, we observe significant performance increases in the EM scores for the EBM-NLP dataset (+8.5 percentage points average across models) in comparison to the other datasets, which we attribute to 
(1) the relatively higher number of Average DS-term mentions per sentence within the dataset,
and (2) the relatively bigger training set size as seen in \autoref{tab:full_dataset_stats}. 
With the exception of NCBI-dataset (with PubMedBERT model), we observe that MSLM achieves consistent performance improvements when detecting full mentions of DS-terms.

\subsection{Is DSFT destructive?}
The success in increasing the sensitivity of LMs towards the DS-terms (via DSFT) is strongly positive as discussed in \textsection\ref{sec:sensitivity_dst}, but at what cost? 
We investigate whether the increased sensitivity comes at the expense of downstream performance, training times and the inherent knowledge of the PLM. For the downstream performance and training times, we monitor the validation NER F1 performance of the MSLM and vanilla flavors over a training time of 20 epochs. \autoref{fig:dsft_f1_scores} shows the MSLM-fine-tuned models consistently outperform the vanilla BioBERT and PubMedBERT during the course of training across the 4 datasets. Furthermore, we observe that MSLM-fine-tuned models achieve the best vanilla performance in a much shorter training time of at most 7 epochs (blue dotted line).

\begin{table}[!t]
\resizebox{\columnwidth}{!}{
\begin{tabular}{@{}lcccc@{}}
\toprule
             & \begin{tabular}[c]{@{}c@{}}BioB\\ (PPL)\end{tabular} & \begin{tabular}[c]{@{}c@{}}BioB\_MSLM\\ (PPL)\end{tabular} & \begin{tabular}[c]{@{}c@{}}Pub\\ (PPL)\end{tabular} & \multicolumn{1}{c}{\begin{tabular}[c]{@{}c@{}}Pub\_MSLM\\ (PPL)\end{tabular}} \\ \midrule
BC2GM        &  1.2          & 2.3 (+1.1)       &  1.2        &   1.9 (+0.7)                                                                           \\
BC5CDR-Chem  &  1.1          & 1.5 (0.4)      & 1.1      &   1.2 (+0.1)                                                                            \\
JNLPBA       & 1.4     &  4.3 (+2.9)       &      1.4  &  3.2 (+1.8)                                                                             \\
NCBI-Disease & 1.2      & 1.3 (+0.1)    &  1.1                                                   &  1.2 (+0.1)                                                                            \\ \bottomrule
\end{tabular}
}
\caption{Validation perplexity (PPL) recorded when the best NER F1 performance was obtained for vanilla and MSLM models.
Biob is BioBERT \& Pub is PubMedBERT, and the change in perplexity when vanilla flavors are replaced by MSLM is indicated in brackets.}
\label{tab:ppl}
\vspace{-1em}
\end{table}

For the inherent knowledge of PLMs, we investigate the validation perplexity to check how well the models understand the domain datasets. As seen in \autoref{tab:ppl}, perplexity increases when MSLM-fine-tuned models replace vanilla models, however, only by a few percentage points. 
We hypothesize that, diminishing the penalties incurred when predicting non DS-terms (as constrained by \eqref{eq:penalty_constraint}) will most likely limit the model's capability to reconstruct corrupted non DS-terms, hence affecting the net perplexity of the models. This change however proves that low perplexity does not necessarily correlate with good performance, a hypothesis also discovered by \citealt{wettig2022should}. 


Overall, the performance improvement achieved by DSFT is evidence supporting the earlier defined hypothesis; i.e. The awareness of or sensitivity of PLMs 
towards DS-terms can be appropriately elevated without hurting downstream performance.

\begin{figure*}[!t]
\centering
\includegraphics[width=\textwidth]{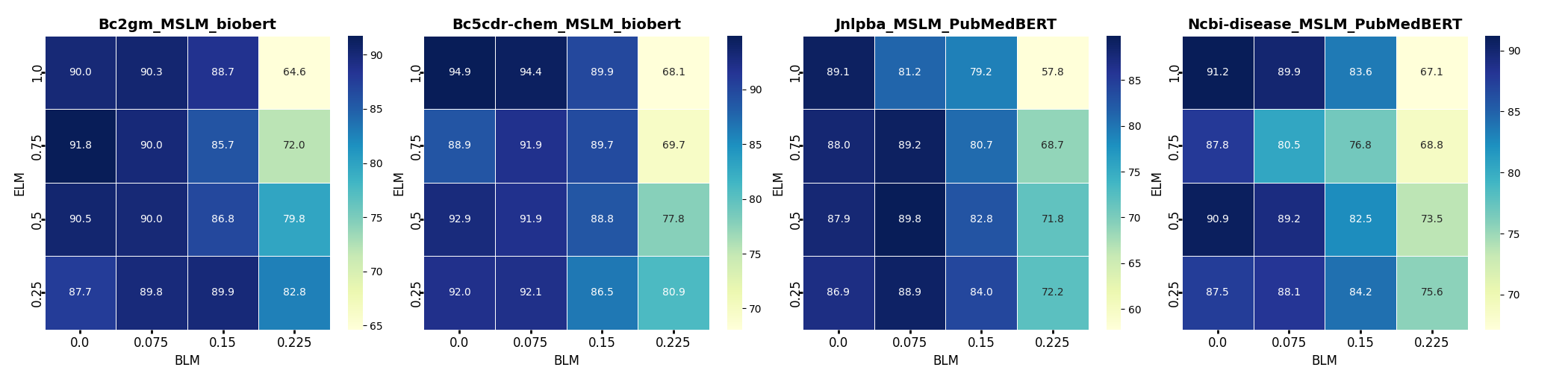}
\caption{Test Exact match (EM) scores of varying ELM and BLM rates when two MSLM-fine-tuned models (MSLM\_biobert and MSLM\_PubMedBERT) are evaluated on the datasets.}
\label{fig:varying_rates}
\vspace{-1em}
\end{figure*}

\section{Analysis}
\subsection{Varying the Masking rates}
\label{sec:varying}
\citealt{devlin2018bert} choose the 15\% masking rate with caution, suggesting that a higher rate risks leaving insufficient context for the LM to learn good representations.  However, this caution can be misleading because, several other factors can influence the optimal masking rates such as the model size and type of the task \cite{liao2020probabilistically}.
We therefore vary the BLM and ELM rates and study the performance changes of the model. To do this, we design the experiments as follows,

\begin{enumerate}
    \item ELM:  We select a range of ELM rates from 25\% to 100\% with interval gaps of 25\%. The interval is kept to 25\% because values < 25\% would not change the overall number of DS-terms to mask, following the ELM mask computation we establish in our setup in \textsection\ref{sec:setup}.
    \item BLM: We select a minimum rate of 0\% and maximum rate of 22.5\% with intervals of 7.5\%. We cap the masking budget for BLM to 22.5\% because we use base models (ca. 125M parameters), which have been reported to struggle in high masking regimes (>20\%) \cite{wettig2022should}. 
    Using a 7.5\% interval is our strategy that enables inclusion of the popular 15\% rate in our set of rates to investigate.
\end{enumerate}
  The resulting sets of rates used in the experiments are [0.25, 0.50, 0.75, 1] and [0, 0.075, 0.15, 0.225] for ELM and BLM respectively.

From \autoref{fig:varying_rates}, we see that increasing both the ELM and BLM rates consistently degrades the performance of the models across all four datasets (i.e. the lowest performance is certainly obtained when both BLM and ELM are high as seen at the top right of all plots). As seen, increasing the BLM rate is only beneficial up to a certain point (7.5\%), and that irrespective of a high or low ELM rate, performance dramatically drops when BLM hits 15\%. These two noticed revelations point to the fact that a high net corruption/masking rate leaves very minimal context to learn from and hence effectively re-construct DS-terms in input sequences, which are already not very long sequences as shown in \autoref{tab:dataset_stats}. Overall, we observe two things, 1) distributing the masking rate budget between the targeted DS-terms and the generic words can contribute to performance gains i.e. optimal scores are obtained when ELM$\ge$0.25 and BLM$\le$0.15, and 2) the optimal Joint ELM-BLM masking rate is dataset dependent as the optimal ELM and BLM rates vary from one dataset to another.

\begin{table}[!t]
\resizebox{\columnwidth}{!}{
\begin{tabular}{@{}cccc@{}}
\toprule
\begin{tabular}[c]{@{}c@{}} \\ ELM{\small (\%)}\end{tabular} & \begin{tabular}[c]{@{}c@{}} \\ BLM{\small (\%)}\end{tabular}  & \begin{tabular}[c]{@{}c@{}}\textless AvgSentLen\\ {[}51{]}\end{tabular} & \begin{tabular}[c]{@{}c@{}}\textgreater AvgSentLen\\ {[}5104{]}\end{tabular} \\ \midrule
100 & 22.5 & 19.4                                                                    & 85.4                                                                         \\
75  & 15.0 & 41.2                                                                    & 84.4                                                                         \\
50  & 7.5  & 75.1                                                                   & 84.0                                                                         \\
25  & 0.0  & 66.3                                                                    & 77.9                                                                         \\ \bottomrule
\end{tabular}
}
\caption{Comparisons of the EM performance of low and high masking regimes for short and long sequences using MSLM\_BioBERT. \textless AvgSenLen {[}51{]} implies, 51 sentences that are shorter than the average sentence length and similarly, \textgreater AvgSentLen, 5104 sentences that are longer than the average sentence length. }
\label{tab:rate_and_length}
\vspace{-1em}
\end{table}

\paragraph{Masking Rate and Sequence Length:}
To further understand how much context is necessary when fine-tuning the MLM, we study the performance of different rates with different sequence lengths on BC2GM.\footnote{We use BC2GM as it has the largest number of sentences below average length compared to the other datasets, which are dominated (ca. 95\%) by sentences above average length} We constrain the rates to low masking regimes, which we define as ELM $\leq$ 0.5 and BLM $\leq$ 0.075, and high masking regimes as ELM $\geq$ 0.75 and BLM $\leq$ 0.15. Because of the laborious nature of the task of constructing a test set with sufficient samples for varying sequence lengths, we use the average sentence length (AvgSentLen in \autoref{tab:dataset_stats}) as a cut off point, where sentences above it are considered as relatively long (>AvgSentLen) and those below as relatively short (<AvgSentLen). We do not perform separate experiments but rather compute the EM scores of the predictions on the short and long sentences. 

In \autoref{tab:rate_and_length}, we observe that high masking regimes favour long sentences (i.e. overall, highest rates produce the best performance for long sentences and worst performance for short ones). This implies that the models are still able to learn sufficiently from long sequences despite a high masking rate. We also observe, while the performance on long sentences is consistently better, it does not significantly differ from that of short ones for the low rates, implying that low rates have minimal impact on varying sequence lengths, and hence LM relies heavily on its inherent pre-trained knowledge.  

\subsection{The effect of mask specific weights}

\begin{table}[!t]
\resizebox{\columnwidth}{!}{
\begin{tabular}{@{}lcccc@{}}
\toprule
                                           & BC2GM & BC5CDR-chem & JNLPBA & NCBI \\ \midrule
BioB\_MSLM                                 & 90.3  & 94.0        & 89.9   & 90.1         \\
$-w_{x \in \{\mathrm{BLM, ELM}\}}$ & 88.7 $\downarrow$  & 93.3 $\downarrow$        & 86.5 $\downarrow$   & 89.9 $\downarrow$        \\ \midrule
Pub\_MSLM                                  & 89.8  & 94.4        & 89.8   & 89.9         \\
$-w_{x \in \{\mathrm{BLM, ELM}\}}$ & 86.9 $\downarrow$  & 94.1 $\downarrow$       & 86.5 $\downarrow$   & 89.1 $\downarrow$        \\ \bottomrule
\end{tabular}
}
\caption{EM scores obtained with and without the mask specific weights. Biob is BioBERT \& Pub is PubMedBERT, $\downarrow$ indicates a performance drop from the originally obtained best scores using the MSLM models.}
\label{tab:ablation}
\end{table}

We perform an ablation analysis to study the impact of the mask specific weights that are used in computing the mask specific losses in the MSLM fine-tuning process (\textsection\ref{sec:msl}). \autoref{tab:ablation} shows that there is performance decline across all experiments when the mask specific weights are eliminated. This performance decline suggests that incorporation of these weights contributes to the performance gains observed in the results.

We attribute these gains to our proposed mask specific weighting scheme which ensures higher weights and hence higher loss costs for the masked named entities (DS-terms) compared to the masked generic words during prediction. Unlike most weighting schemes that consider the overall class distribution in the data, the weighting scheme considers the distribution of masked units rather than of classes. Furthermore, the scheme mitigates against overfitting the LM’s weights to DS-terms by introducing a sensitivity threshold, which is used to encourage the weights of DS-terms while carefully suppressing weights of generic terms. For instance, if eq \eqref{eq:distribution_weight} provides $w_{\mathrm{ELM}}=0.4$, eq \eqref{eq:penalty_constraint} will calibrate the weight giving $w_{\mathrm{ELM}}=\max(0.5, 0.4)=0.5$, implying the weight for DS-terms should never fall below $0.5$, and similarly if eq \eqref{eq:distribution_weight} provides $w_{\mathrm{BLM}} = 0.6$, eq \eqref{eq:penalty_constraint} will calibrate the weight as $w_{\mathrm{BLM}} = \min(0.5, 0.6) = 0.5$, implying that the weight for generic or random words should never rise above $0.5$. The intuition is that the model becomes more sensitive to DS-terms but also keeping it aware of the context surrounding the DS-terms.

\begin{figure}[!b]
\centering
\includegraphics[width=1\columnwidth]{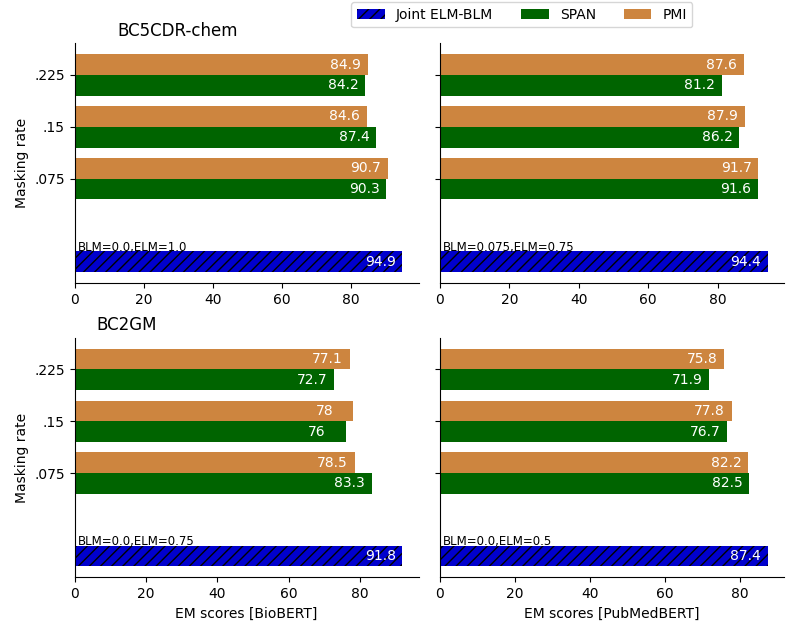}
\caption{Comparing performance of other masking strategies across various rates with the best performance of our proposed Joint ELM-BLM. Results of BioBERT (left) and PubMedBERT (right) evaluated on BC2GM and BC5CDR-chem and hatch the bars with best scores.}
\label{fig:varying_strategies}
\vspace{-1em}
\end{figure}

\section{Comparisons with Prior Masking Strategies}
Besides our proposed masking strategy (i.e. Joint ELM-BLM), there are various other advanced masking strategies such as PMI-Masking (PMI) \cite{levine2020pmi} and Random-Span masking (SPAN) \cite{joshi2020spanbert}. With PMI, spans of co-occurring words (2-4) (a.k.a collocations) are identified, ranked based on PMI scores computed using the PMI measure proposed by \newcite{levine2020pmi} and stored in a vocabulary. The ranked spans discovered in an input sequence are masked. In the SPAN approach, spans of varying lengths (2-4) are arbitrarily selected and masked. In both approaches, the total masking budget (number of tokens to mask) is maintained to avoid biasing the comparative analysis. Extended details of how we implement SPAN and PMI are in Appendix \ref{sec:appendix_strategies}.

In \autoref{fig:varying_strategies}, we directly replace Joint ELM-BLM with either SPAN or PMI and vary the total masking budget while maintaining the optimal budget for Joint ELM-BLM. For instance, we keep ELM = 0.75 and BLM = 0.0 for Joint ELM-BLM, when evaluating MSLM\_Biobert on BC2GM dataset because they are the optimal rates. However, we vary the rates for both PMI and SPAN across the values in the set of BLM rates established in \textsection\ref{sec:varying}. 

We observe that Joint ELM-BLM outperforms other strategies across all experiments. PMI produces majority of the second best results despite SPAN masking being quite competitive. 
We attribute PMI's performance to the fact that the PMI's vocabulary from which spans to mask are drawn has a high concentration (> 50\%) of DS-terms (details in Appendix \ref{sec:overlap_appendix})), which effectively makes it similar to ELM that directly masks DS-terms. As noticed earlier in \textsection\ref{sec:varying}, masking DS-terms is highy effective even with no BLM masking (i.e. BLM = 0.0). 
We also observe the slight drop in performance as the masking rate increases across PMI and SPAN, which further confirms the fact that LMs are likely to struggle when decoding highly corrupted sequences \cite{devlin2018bert}.




\section{Related work}

\textbf{Domain adaptation of PLMs for NER:}
The conventional approach in prior work tackling domain adaptation for NER has focused  pre-training on unlabelled target domain corpora and then fine-tune on downstream target domain dataset \cite{lee2020biobert, beltagy2019scibert}. Recent work has explored minimising the discrepancy between the source and target embedding distributions \cite{zhang2021pdaln, poerner2020inexpensive}. Our work mostly aligns with \newcite{poerner2020inexpensive} who also adopt ``non-target domain pre-training''.
\\ 

\begingroup
\raggedright 
\textbf{Masking:} The originally proposed masking approach that involved replacing a percentage of tokens at random (TOKEN masking) with $\mathrm{[MASK]}$ tokens \cite{devlin2018bert} has been modified in recent works to improve MLM.  \newcite{sun2019ernie} and \newcite{abaho2022position} mask named entity spans (entity masking), \newcite{joshi2020spanbert} mask random spans of tokens (SPAN masking) and \newcite{levine2020pmi} mask groups of co-occurring words (PMI masking).
\endgroup
With the exception of PMI, our proposed Joint ELM-BLM masking approach aligns well with all recent masking modifications. It simultaneously masks disjoint sets of random tokens and entity spans. Targeting multiple units in a sentence makes it greedier than prior works, however, we emphasize mask rate tuning and upholding a masking budget to achieve optimal performance. 

\section{Conclusion}
We considered the problem of DSFT aiming to improve an LM's sensitivity (i.e. awareness of) towards DS-terms. We proposed MSLM, an approach that jointly masks DS-terms and random words, while conditioning the LM to larger penalties during optimisation for incorrect predictions of DS-terms. Using the biomedical domain as a testbed, the performed experiments reveal improvements MSLM makes over vanilla fine-tuning in exact DS-term match detection. MSLM's efficiency is proven when models achieve higher NER F1 scores in a much shorter training time. We substantiate the recent narrative, dismissing 15\% as a universally optimal rate in MLM \cite{wettig2022should}, by proving that optimal performance is influenced by varying masking rates and length of sequences. 

The Joint ELM-BLM masking strategy we propose outperforms advanced masking methods. Although we focus on biomedical NER, our proposed MSLM approach can be be adapted for DSFT for other domains.
The positive impact of our proposed masking method motivates us to investigate its effectiveness during pre-training of MLMs in future work.

\section*{Limitations}
The list of pre-trained biomedical LMs we use in our experiments can be considered as a representative sample that is used frequently for biomedical text mining. However, there are some other biomedical LMs such ClinicalBERT \cite{alsentzer2019publicly} and BlueBERT \cite{, peng2020empirical}, whose inclusion can quantitatively improve results of our analysis.  
Despite casting it as an NER task focused on not simply detecting DS-terms, but confidently detecting them for that matter, some other tasks worthy of consideration for investigating sensitivity may include but not limited to, question and answering \cite{choi2018quac}, common sense reasoning \cite{davis2015commonsense}, event detection \cite{weng2011event} etc. Furthermore, studying the performance of domain sensitive fine-tuning in other domains besides biomedicine would be a qualitative addition and is recommendable for future research under the guise of improving LM sensitivity. 

\section*{Ethics}
This work addresses insensitive fine-tuning that arises from the neglection of the disparities and nuances between source and target domains. In addressing this problem, our proposed fine-tuning method neither guards against nor removes any present biases (social, gender etc) in the pre-trained MLMs. 

Additionally, we do not annotate any data for the datasets we adopt as they are all existing datasets within the BLURB benchmark \cite{gu2021domain} that are commonly used for biomedical text mining. 

Furthermore, we credit all prior work whose output directly or indirectly influences our work especially with the datasets and the methods. In our evaluation experiments, we declare some results that were not generated from a seperate set of experiments but instead obtained by selectively retrieving a set of sentences that conform to the evaluation criteria we targeted i.e. short and long sentences. In comparing our masking strategy to the advanced bench-marking strategies, we study performance across various masking budgets in order to provide a fair comparison with our proposed method.  To further remove any modelling bias, we elaborately discuss implementation details of compared methods in Appendix.

\section*{Acknowledgements}
We wish to acknowledge funding by the \href{https://www.nihr.ac.uk/}{NIHR} for the \href{https://www.liverpool.ac.uk/dynairx/}{DynAIRx} and \href{https://mric.uk/}{MRIC} projects hosted by the Civic Health Innovation Labs (\href{https://www.liverpool.ac.uk/civic-health-innovation-labs/}{CHIL}). These health informatics projects inspired the research problems addressed in this work. We specifically thank the wider DynAIRx team for the valuable discussions that shaped the ideas and propositions made in this work. We are also thankful to the reviewing team whose feedback was necessary in improving the work in the paper.   

\bibliography{custom}
\bibliographystyle{acl_natbib}

\section*{Appendices}
\appendix

\section{Dataset statistics}
\label{sec:appendix_stats}
The full table containing dataset statistics partially presented in \autoref{tab:dataset_stats}, is shown in \autoref{tab:full_dataset_stats}. 

\section{Hyperparameters}
\begin{table}[!h]
\centering
\resizebox{\columnwidth}{!}{
\begin{tabular}{@{}lcc@{}}
\toprule
\textbf{Parameter}                                                             & \textbf{Tuned-range}                                                                & \textbf{Optimal}                                   \\ \midrule
Train Batch size                                                                     & {[}8,16,32{]}                                                                      & 8                                                 \\
Eval Batch size                                                                     & {[}8,16,32{]}                                                                      & 8                                                \\
Epochs                                                                       & {[}10,20,30,50{]}                                                           & 20                                              \\
$k$                                                                      & {[}50, 100,200,300{]}                                                           & 100                                              \\
$d$                                                                       & {[}50,100,200,300{]}                                                           & 100                                             \\
Optimizer                                                                      & {[}Adam, SGD{]}                                                                     & Adam                                               \\

Learning rate                                                                  & \begin{tabular}[c]{@{}c@{}}{[}5e-5, 1e-4, 5e-3, 1e-3{]}\end{tabular} & 5e-5                                               \\ \bottomrule
\end{tabular}
}
\caption{Parameter settings for the MSLM-fine-tuned models. $k$ and $d$ are dimensions of the the randomly initialised trainable weight vectors $\vec{W}^{(ed)} \in \R^{1 \times k}$ defined in  \ref{eq:entity_detection} and $\vec{W}^{(ec)} \in \R^{1 \times d}$ defined in \ref{eq:entity_classification} respectively.}
\label{tab:parameter_pbc}
\vspace{-1em}
\end{table}

\begin{table}[!t]
\resizebox{\columnwidth}{!}{
\begin{tabular}{@{}llcc@{}}
\toprule
             &            & \textbf{Vanilla}  & \begin{tabular}[c]{@{}c@{}}\textbf{MSLM}\\ {\small BLM=0.075 ELM=1}\end{tabular} \\ \midrule
BC2GM        & BioBERT    &   88.4           &   \textbf{90.3$_{\pm 0.5}$}                                                                    \\
             & PubMedBERT &  86.8           & \textbf{89.8$_{\pm 0.4}$}                                                                      \\
             & BioELECTRA &  87.6               &   89.1$_{\pm 0.2}$                                                                      \\
             & SciBERT &  85.7              &  \textbf{87.1$_{\pm 0.4}$}   
                                                         \\
NCBI-disease & BioBERT    &    89.1         &     \textbf{90.1$_{\pm 0.1}$}                                                                  \\
             & PubMedBERT &  \textbf{89.9}           &  \textbf{89.9$_{\pm 0.2}$}                                                                      \\
             & BioELECTRA &  88.5               &    \textbf{88.9$_{\pm 0.2}$}                                                                    \\
             & SciBERT &  88.4              & \textbf{89.9$_{\pm 0.1}$}    
                                                         \\

BC5DCR-chem      & BioBERT    &   93.3      &  \textbf{94.0$_{\pm 0.2}$}                                                                      \\
             & PubMedBERT &  94.0           & \textbf{94.4$_{\pm 0.2}$}                                                                        \\
             & BioELECTRA &  90.8              &  \textbf{94.0$_{\pm 0.2}$}                                                                  \\
             & SciBERT &   90.7               & \textbf{93.7$_{\pm 0.2}$}   
                                                         \\

EBM-NLP      & BioBERT    &  64.3        &   \textbf{75.4$_{\pm 0.4}$}                                                                    \\
             & PubMedBERT &  65.5        &   \textbf{76.2$_{\pm 0.3}$}                                                                    \\
             & BioELECTRA &  63.7            &   \textbf{73.2$_{\pm 0.3}$}                                                                       \\
             & SciBERT &  69.7      &   \textbf{73.4 $_{\pm 0.2}$} 
                                                         \\
BC5DCR-dis      & BioBERT    &   91.7      &   \textbf{93.4 $_{\pm 0.2}$}                                                                \\
             & PubMedBERT &   92.3   &    \textbf{94.1 $_{\pm 0.1}$}                                                                 \\
             & BioELECTRA &   89.7        & \textbf{93.5 $_{\pm 0.3}$}                                                                       \\
             & SciBERT &   90.1          &  \textbf{93.4 $_{\pm 0.2}$}   
                                                         \\
JNLPBA      & BioBERT    &   86.3   &        \textbf{88.9 $_{\pm 0.2}$}                                                               \\
             & PubMedBERT &   85.7        &      \textbf{89.8 $_{\pm 0.2}$}                                                                \\
             & BioELECTRA &  80.0            &   \textbf{83.4 $_{\pm 0.2}$}                                                                      \\
             & SciBERT &   82.4          &   \textbf{85.4 $_{\pm 0.2}$} 
                                                         \\
                                                         \bottomrule
\end{tabular}
}
\caption{Full Exact match scores obtained when MSLM is initialized with various pre-trained biomedical LMs. These scores are compared against Vanilla versions of the LMs. Best and second-best are bold and underlined. Partial results of the table are presented in the main body in \autoref{tab:exact_match_scores}.}
\label{tab:full_exact_match_scores}
\vspace{-1em}
\end{table}

\begin{table*}[t]
\centering
\resizebox{\textwidth}{!}{
\begin{tabular}{@{}lcccccc@{}}
\toprule
 & \begin{tabular}[c]{@{}c@{}}\#Sents\\ {\small Train | Val | Test}\end{tabular} & \#Classes & AvgSentLen & \multicolumn{1}{c}{\begin{tabular}[c]{@{}c@{}}\#Ments\\ {\small Train | Val | Test}\end{tabular}} & AvgMents & AvgMentsLen \\ \midrule
BC2GM &        12632 | 2531 | 5065    & 2          &  25.17          & 15197 | 3061 | 632  & 1.20 & 2.4 \\
NCBI-disease &  5432 | 923 | 942      & 2          &  25.24          & 5134 | 787 | 960    & 0.95 & 2.2 \\  
BC5CDR-chem &  4812 | 4602 | 4582     & 2          &  25.75          & 5385 | 5203 | 5347  & 1.12 & 1.3\\ 
BC5CDR-dis &  4812 | 4602 | 4582      & 2          &  25.75          & 4182 | 4246 | 4424  & 0.87 & 1.7 \\
JNLPBA &  14731 | 3876 | 3873         & 2          &  30.05          & 32178 | 8575 | 6241 & 2.18 & 3.0 \\ 
EBM-NLP &  32074 | 4009 | 4010        & 5          &  24.68          & 21498 | 2677 | 2736 & 2.67 & 2.0 \\ 
MIMIC III &  9937 | 1242 | 1243        & 3          &  1943.85          & 863732 | 106539 | 107330 & 8.67 & 2.0 \\ \bottomrule
\end{tabular}
}
\caption{Dataset statistics. \#Sents and \#Ments are the number of sentences and number of DS-term mentions respectively for the train, validation and test splits, AvgSentLen is the Average length of sentences, AvgMents is the Average number of DS-terms mentioned per sentence obtained as (\# of train Ent\_Ments)/(\# of train sents) and AvgMentsLen is the average length of DS-terms.}
\label{tab:full_dataset_stats}
\vspace{-1em}
\end{table*}

\section{Sensitivity towards DS-terms}

\autoref{tab:full_exact_match_scores} presents the full results of EM scores in detecting full or exact mentions of DS-terms. We observe an average increment of +3.2 points across all datasets when all four LMs are used.

\begin{table}[!t]
\resizebox{\columnwidth}{!}{
\begin{tabular}{@{}lccc@{}}
\toprule
               & \begin{tabular}[c]{@{}c@{}}BioBERT\\ {\small \cite{lee2020biobert}}\end{tabular} & \begin{tabular}[c]{@{}c@{}}GreenBioBERT\\ {\small \cite{poerner2020inexpensive}}\end{tabular} & \begin{tabular}[c]{@{}c@{}}MSLM-BioBERT\\ {\small ELM=1,BLM=0.075}\end{tabular} \\ \midrule
BC5CDR-disease & \underline{87.15}                                                 & 85.08                                                            & \textbf{89.45}                                                         \\
NCBI-disease   & \underline{89.71}                                                 & 85.94                                                            & \textbf{91.91}                                                         \\
BC5CDR-chem    & \underline{93.47}                                                 & 93.08                                                            & \textbf{96.79}                                                         \\
BC2GM          & \underline{84.72}                                                 & 83.45                                                            & \textbf{92.17}                                                         \\
JNLPBA         & \underline{77.49}                                                 & 76.89                                                            & \textbf{83.24}                                                         \\ \bottomrule
\end{tabular}
}
\caption{Downstream NER test F1 scores when different variants of BioBERT are fine-tuned on the datasets. Reference scores from compared methods \cite{lee2020biobert} and \cite{poerner2020inexpensive}. Best and second best results are in bold and underlined respectively.}
\label{tab:F1}
\end{table}

\section{Domain Adaptation}
Even if we do not technically have a source and target domain for respectively pre-training fine-tuning, our work aligns with prior work which achieves domain adaptation without pre-training on a target domain. \citealt{poerner2020inexpensive} build a model called greenBioBERT in a relatively less expensive approach and fine-tune it on the same datasets we do. greenBioBERT is word2vec trained on PubMed+PMC articles and with an updated embedding layer and tokenizer following BERT's architecture. The authors consider this as an LM not pre-trained on target domain.

We compare test NER F1 perfomance in our experiments with both grteenBioBERT and vanilla BioBERT. Results in \autoref{tab:F1} show our MSLM-fine-tuned BioBERT outperform all the others by at least +2.3 points. This further indicates the heitened awareness of DS-terms that MSLM is able to achieve hence effectively improving its entity detection performance. 

\begin{algorithm}[!htb]
\caption{SPAN Masking}
\label{alg:algorithm_span}
\begin{algorithmic}[1] 
\STATE \textbf{Input}: Tokenized input sequence:- $s$, \\ masking\_rate:- $m_r$, mask token:- $\mathrm{[MASK]}$, \\ 
\textbf{Output}: Masked Tokenized Input sequence $s_{\mathrm{M}}$

\STATE Initialize the below, \\
  \hspace{0.8em} - A pool of indices ($s_{\textrm{random\_pool}}$) randomly
  \hspace{0.8em} ordered, where $|s|$ = $|s_{\textrm{random\_pool}}|$ \\
  \hspace{0.8em} - masked\_budget $mb$ = $\mathrm{math.ceil}(m_r \times |s|$) \\ 
  \hspace{0.8em} - masked\_so\_far $msf = 0$

\FOR{index $i$ in $s_{\textrm{random\_pool}}$}
    \STATE Initialize random\_span\_length $sl_{i=2}^{4}$ \\ i.e. span to be masked could vary from length 2 to 4.
    \STATE $sl$ = $\min(sl, mb)$
    \STATE start, end = $i$, $i$+$sl$ \\
    {\small Don't mask beyond the masking budget [7-10]}
    \IF{$(msf + sl) > mb$:} 
        \STATE $sl$ = $mb$ - $msf$
        \STATE end = $i$+$sl$
    \ENDIF
    \\
    {\small Don't mask beyond sequence bounds [11-13]}
    \IF{$end \ge |s| - 1$} 
        \STATE end = $i+sl$
        \STATE $sl$ = end - start
    \ENDIF
    \\
    {\small Don't mask already masked spans [15-17]}
    \IF{$s_{\mathrm{M}}[$start:end$]$ has no $\mathrm{[MASK]}$ tokens} 
        \STATE $s_{\mathrm{M}}[start$:$end]$ = $\mathrm{[MASK]} * sl$
        \STATE $\mathrm{msf}$ += $sl$
    \ENDIF

    \IF{$\mathrm{msf} \ge \mathrm{mb}$} 
        \STATE break
    \ENDIF
\ENDFOR 
\hspace{1em} return $s_{\mathrm{M}}$
\end{algorithmic}
\end{algorithm}

\section{Masking strategies}
\label{sec:appendix_strategies}
We compare our proposed joint ELM-BLM masking strategy to two other advanced masking strategies, PMI \cite{levine2020pmi} and Random SPAN \cite{joshi2020spanbert} whose implementation we respectively present in a pseudo code in the algorithms \ref{alg:algorithm_pmi} and \ref{alg:algorithm_span}.

\subsection{SPAN Masking (\ref{alg:algorithm_span})}
Given a tokenized input sequence and a masking rate $m_r$ as input (line 1), we initialize a pool of indices (of the same size as the input sequence $|s|$) randomly ordered ($s_{\mathrm{random\_pool}}$). Each random index is a possible starting index of a contiguous span to be masked. We compute the masking budget $m_b$ as product between rate and input sequence size to get number of tokens to be masked e.g. if $|s| = 10$ and $m_r = 0.15$, $mb = 0.15 \times 10$. For each random index in the pool $s_{random_{pool}}$, we initialize a span length $s_l$ randomly $s_l \in {2,3,4}$ at line 4 i.e. this is the length of the contiguous span to be masked.
Three different constraints satisfied as we iteratively select random spans to be masked include, 1) the number of already masked tokens summed up with span length $s_l$ should be less than the masking budget $m_b$ (line 7-9), 2) then the end index of span to masked should not be greater than the end index of the input sequence (line 11-13), then finally the selected span to be masked should not contain already masked tokens inhibiting overlapping masking (line 15-17). Once all constraints are satisfied, the span's tokens within the input sequence are masked or replaced with mask token $\mathrm{[MASK]}$.

\begin{algorithm}[!b]
\caption{PMI Masking}
\label{alg:algorithm_pmi}
\begin{algorithmic}[1] 
\STATE \textbf{Input}: Tokenized input sequence:- $s$, \\ masking\_rate:- $m_r$, mask token:- $\mathrm{[MASK]}$, PMI\_vocabularly (PMI$_{v}$) \\ 
\textbf{Output}: Masked Tokenized Input sequence $s_{\mathrm{M}}$
\STATE Initialize the below, \\
  \hspace{0.8em} - masked\_budget $mb = \mathrm{math.ceil}(m_r \times |s|$) \\ 
  \hspace{0.8em} - masked\_so\_far $msf = 0$ \\
\WHILE{$msf \le mb$}
  \FOR{gram in PMI$_v$}
    \IF{gram is a subsequence in $s_{\mathrm{M}}$} 
        \STATE Get start (st) and end (ed) indices of gram in $s_{\mathrm{M}}$
        \STATE gram$_l$ = |gram|
        \IF{$msf$ + gram$_l$ $> mb$} 
            \STATE gram$_l$ = $mb$ - $msf$
            \STATE end = $st$+gram$_l$
        \ENDIF
        \STATE $s_{\mathrm{M}}[st$:$ed]$ = $\mathrm{[MASK]} *$ gram$_l$
        \STATE $\mathrm{msf}$ += gram$_l$
    \ENDIF
  \ENDFOR
\ENDWHILE
\STATE return $s_{\mathrm{M}}$
\end{algorithmic}
\end{algorithm}

\subsection{PMI Masking (\ref{alg:algorithm_pmi})}
With PMI, we begin by constructing a PMI vocabulary of word n-grams of lengths 2–4. These n-grams contain words that co-occur in sentences a minimum of 5 times within the entire dataset. A PMI score for each collocation (n-gram of co-occurying words) is computed using the PMI measure \cite{levine2020pmi}. The collocations are ranked and ordered in their respective lengths. \\
NB: Each dataset has its own PMI vocabulary.

Given a tokenized input sequence and a masking rate $m_r$ as input (line 1). The masking budget $m_b$ is computed similar to the SPAN approach (line 2). For each collocation (gram) in the vocabulary, we check if collocation is a subsequence (contiguous) of the input sequence. One constraint satisfied is 1) the number of already masked tokens summed up with span length $s_l$ should be less than the masking budget $m_b$ (line 8-10), 2). Once constraint is satisfied, the span's tokens within the input sequence are masked or replaced with mask token $\mathrm{[MASK]}$. 

\subsection{PMI vocabularly overlapping DS-terms}
\label{sec:overlap_appendix}
\begin{table}[!htb]
\centering
\begin{tabular}{@{}ccc@{}}
\toprule
\#DS-terms & \#PMI-vocab & \begin{tabular}[c]{@{}c@{}}\#Overlap\\ (\# | \%)\end{tabular} \\ \midrule
18890      & 15787       & 8130 | 51.5                                                     \\ \bottomrule
\end{tabular}
\caption{Number of vocabularly terms that overlap across with DS-terms in the BC2GM dataset. ``\#'' implies number of, \% implies percentage of the vocabularly that are DS-terms.}
\label{tab:overlapping_pmi}
\end{table}

\autoref{tab:overlapping_pmi} shows that 51.5\% of the phrases in the constructed PMI's vocabularly (for the BC2GM dataset) are DS-terms. This high concentration of DS-terms in the PMI vocabularly implies that there is a high similarity between PMI masking and Entity Level Masking (ELM) and hence making PMI masking nearly as effective as standalone ELM masking (i.e. even without BLM masking). \autoref{tab:pmi_vocab} shows a sample of the DS-terms that overlap (in blue) across with the PMI vocabularly. 

\begin{figure*}[!t]
\centering
\includegraphics[width=0.9\textwidth, height=0.7\textwidth]{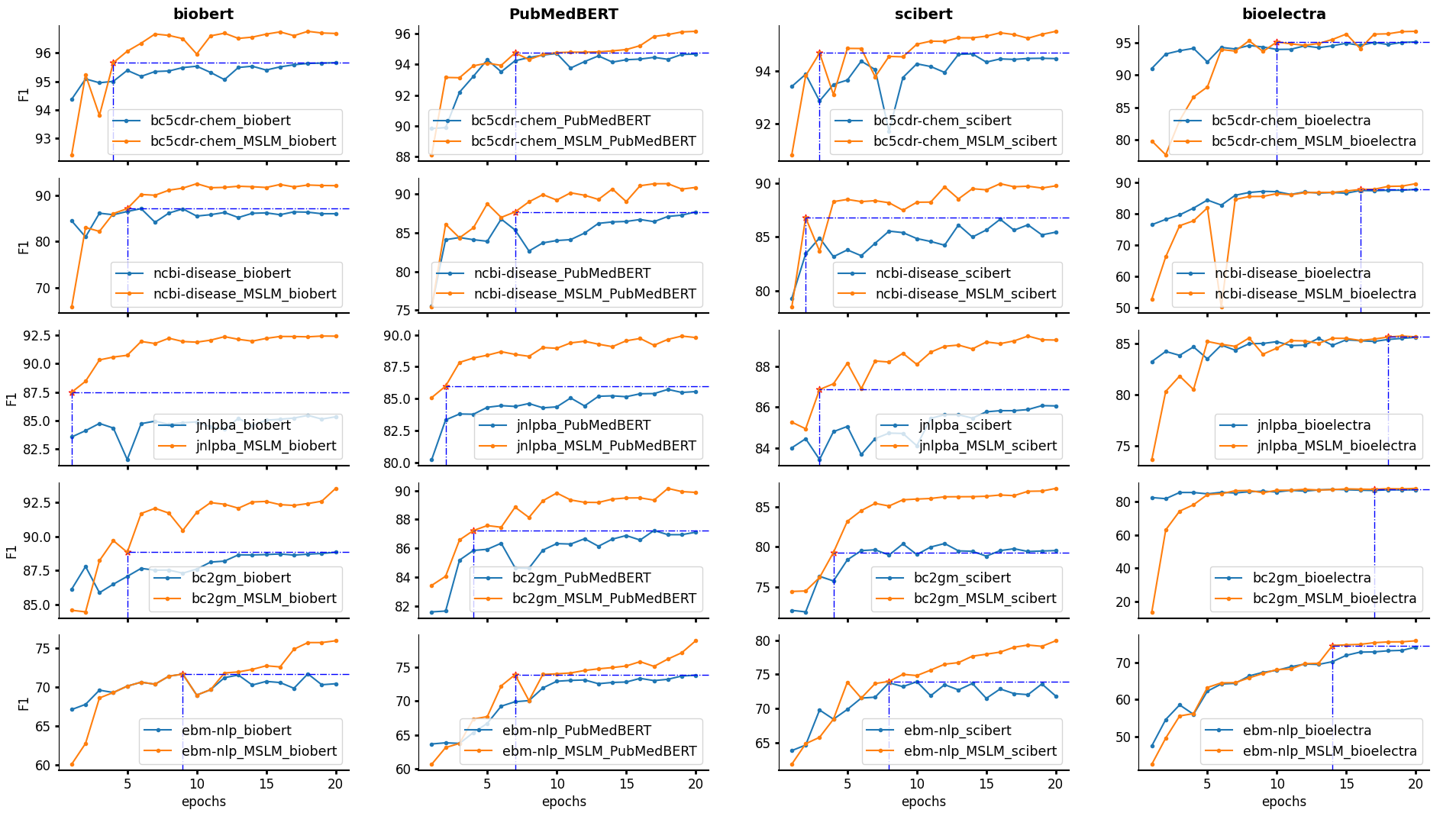}
\caption{Downstream NER F1 performance of the vanilla and the MSLM-fine-tuned models. "MSLM" is used to uniquely identify DSFT models. ELM and BLM rates used in §3.1 are maintained. Each row contains results specific to a dataset e.g. first row has BC5CDR-chem, second has NCBI-disease etc. Similarly each column contains results specific to pre-trained biomedical LM.}
\label{fig:all_f1}
\vspace{-1em}
\end{figure*}

\section{Is DSFT destructive?}
\label{sec:dsft_destructive}
We present the complete list of all plots from the experiments investigating whether DSFT is destructive hence exploring an answer to the hypothesis in the introduction, i.e. the awareness of or sensitivity towards DS-terms can be appropriately elevated when fine-tuning without hurting downstream performance.

As observed in \autoref{fig:all_f1}, we observe better results achieved by the MSLM fine-tuned models, more so, achieving the best performance of the vanilla models in a much shorter training time. A couple of other things we notive include, performance during the course of training of bioelectra models doesn't seem to signigicantly differ from that of the MSLM\_bioelectra models across all datasets. We also notice that unlike all the other models, with bioelectra, MSLM\_fine-tuned models achieve the best performance of the vanilla models after 10 epochs, i.e. longer than the other models. We attribute bioelectra's competitiveness to its inherent architectures~\cite[ELECTRA;][]{clark2020electra} which, similar to MSLM, it adds a model to detect whether MLM has correctly replaced a token or not (token replacement detection). Electra trains a generator (which is an MLM) to predict tokens for masked slots, and additionally trains a discriminator to predict whether a token has been replaced or the original masked token is what the generator predicted. Whereas MSLM doesn't add any model on top of the MLM, it targets MLM components i.e. tilting the MLMs sensitivity towards masks tokens corresponding to DS-terms. 

\begin{figure*}
\centering
\includegraphics[width=0.80\textwidth]{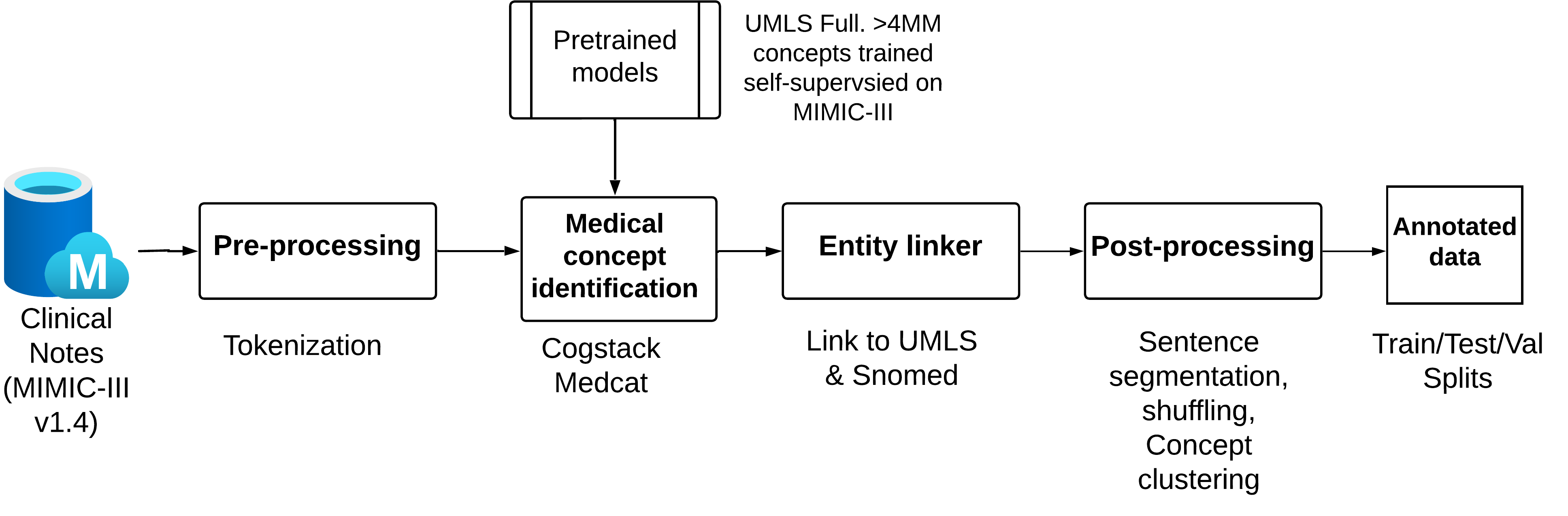}
\caption{Weakly supervised annotation of MIMIC-III v1.4.}
\label{fig:weak_supervised}
\end{figure*}

\section{Additional Analysis}
Due to space limitations, we defer additional investigations to further validate our MSLM approach to this Appendix. We investigate MSLM in a weakly supervised setting and detail everything in following sections.


\subsection{Weak supervision of MIMIC-III}
\label{sec:appendix_weak_supervised}
Specifically, we employ \href{https://physionet.org/content/mimiciii/1.4/}{MIMIC-III v1.4} \cite{johnson2016mimic} dataset, and retrieve a sample of 5000 patient records from the NOTESEVENT table (within the MIMIC-III v1.4 database) containing de-identified free text entries recorded by physicians and other care providers during patient-care. \autoref{fig:weak_supervised} illustrates the pipeline used in annotating mimic-III in a weakly supervised process. 

We use Cogstack medcat\footnote{\url{https://medcat.readthedocs.io/en/latest/index.html}},
a biomedical annotation tool, to extract and categorise medical concepts based on medical semantic types defined\footnote{\url{https://lhncbc.nlm.nih.gov/ii/tools/MetaMap/documentation/SemanticTypesAndGroups.html}} in UMLS and Snomed. 

\begin{table}[!t]
\resizebox{\columnwidth}{!}{
\begin{tabular}{ll}
\hline
Cluster Category & Associated UMLS Semantic Types                                                                                                                                                                                                                                                                   \\ \hline \toprule
Treatments        & \begin{tabular}[c]{@{}l@{}}{[}"Pharmacologic Substance", "Clinical Drug",\\ Antibiotic{]}\end{tabular}                                                                                                                                                                                           \\ \midrule
Diseases         & \begin{tabular}[c]{@{}l@{}}{[}"Acquired Abnormality", \\ "Anatomical Abnormality", "Bacterium", \\ "Archaeon", "Congenital Abnormality",           \\ "Cell or Molecular Dysfunction", \\ "Disease or Syndrome", "Virus",        \\ "Neoplastic process"{]}\end{tabular} \\ \midrule
Symptoms          & \begin{tabular}[c]{@{}l@{}}{[}"Social Behavior", "Sign or Symptom", \\ "Mental or Behavioral Dysfunction"{]}\end{tabular}                                                                                                                                                                        \\ \hline
\end{tabular}
}
\caption{UMLS semantic types that Cogstack can link to are clustered into three high level categories by a clinical consultant. These clusters encapsulate the semantic types in an easy-to understand manner}
\label{tab:semantic_type_cluster}
\vspace{-1em}
\end{table}

Because of the unequal distribution of the semantic types across the annotations, we narrow down the scope of target UMLS semantic concepts with the help of a clinical consultant
who clusters concepts into three high-level clinical
concepts of Diseases, Symptoms and Treatments, as shown in \autoref{tab:semantic_type_cluster}.

After the annotations, we then use SpaCy\footnote{\url{https://spacy.io/}} \cite{neumann2019scispacy} for sentence segmentation of each record (a row containing multiple paragraphs) and split the resulting list of sentences into train, validation and test sets (9937, 1242 and 1243 sentences respectively), which are then subsequently used in fine-tuning.

\subsection{Results}
After preliminary tuning of BLM and ELM rates on validation set, we find the optimal BLM and ELM rate as 0.075 and 0.5 respectively, which achieves an average improvement of 2.7 points in EM scores over the vanilla approach as seen in \autoref{tab:mimic_exact_match_scores}. This improvement further indicates how beneficial MSLM is in improving extraction of DS-terms from clinical patient data. 
rather than just scientific literature in BLURB datasets.

\begin{table}[!t]
\resizebox{\columnwidth}{!}{
\begin{tabular}{@{}llccc@{}}
\toprule
             &            & \textbf{Vanilla}  & \begin{tabular}[c]{@{}c@{}}\textbf{MSLM}\\ {\small BLM=0.075 ELM=0.5}\end{tabular} \\ \midrule
MIMIC-III    & BioBERT    &   90.1           &   \textbf{92.6$_{\pm 0.2}$}                                                                    \\
             & PubMedBERT &  89.8           & \textbf{93.8$_{\pm 0.4}$}                                                                      \\
             & BioELECTRA &  88.1               &   \textbf{90.1$_{\pm 0.2}$}                                                                      \\
             & SciBERT &  87.5              &  \textbf{89.7$_{\pm 0.4}$}    
                                                         \\ \bottomrule
\end{tabular}
}
\caption{Exact match (EM) scores. Average scores
(across 5 runs) obtained for fine-tuning LMs on weakly supervised dataset constructed using MIMIC-III patient records.}
\label{tab:mimic_exact_match_scores}
\vspace{-1em}
\end{table}

\begin{table*}[]
\resizebox{\textwidth}{!}{
\begin{tabular}{@{}lll@{}}
\toprule
\multicolumn{3}{c}{BC2GM PMI Vocabulary}                                                                                                                                          \\ \midrule
 \begin{tabular}[c]{@{}l@{}}
Bombyx mori
 \\ 
\textbf{\textcolor{blue}{IE promoter}} \\ 
CASE REPORT
 \\ 
Codonopsis pilosula
 \\ 
\textbf{\textcolor{blue}{E2 proteins}} \\ 
\textbf{\textcolor{blue}{HMR locus}} \\ 
LY 294002
 \\ 
Leptomonas seymouri
 \\ 
OAE screener
 \\ 
\textbf{\textcolor{blue}{latent membrane protein 2A}} \\ 
Pisum sativum
 \\ 
\textbf{\textcolor{blue}{protein tyrosine kinase}} \\ 
Punta Toro
 \\ 
Rhodosporidium toruloides
 \\ 
\textbf{\textcolor{blue}{PDH complex}} \\ 
\textbf{\textcolor{blue}{dopamine D2 receptor}} \\ 
Trait Personality
 \\ 
Van der
 \\ 
Veterans Affairs
 \\ 
\textbf{\textcolor{blue}{human chorionic gonadotropin}} \\ 
chengchi tang
 \\ 
\textbf{\textcolor{blue}{cysteine proteinase}} \\ 
dig1 dig2
 \\ 
dihydrolipoyl transsuccinylase
 \\ 
\textbf{\textcolor{blue}{bacterial chloramphenicol acetyltransferase}} \\ 
\textbf{\textcolor{blue}{chloroacetate esterase}} \\ 
{\centering{
$\cdot$}}
 \\ 
{\centering{
$\cdot$}}
 \\ 
{\centering{
$\cdot$}}
 \\ 
\textbf{\textcolor{blue}{bHLH proteins}} \\ 
ta chengchi
 \\ 
\textbf{\textcolor{blue}{pleckstrin homology domain}} \\ 
Aedes aegypti
 \\ 
Autographa californica
 \\ 
\textbf{\textcolor{blue}{RNAP II}} \\ 
\textbf{\textcolor{blue}{sigma 54}} \\ 
El Paso
 \\ 
Expiratory Flow
 \\ 
Gulf War
 \\ 
\textbf{\textcolor{blue}{Hematopoietic growth factors}} \\ 
Rhodobacter capsulatus
 \\ 
\textbf{\textcolor{blue}{Src homology}} \\ 
Task Force
 \\ 
Toxocara canis
 \\ 
 \textbf{\textcolor{blue}{monoamine oxidase}} \\ 
\textbf{\textcolor{blue}{cytochrome oxidase}} \\ 
acne vulgaris
 \\ 
aluminium hydroxide
 \\ 
binocular pregeniculate
 \\ 
\textbf{\textcolor{blue}{U5 RNA}} \\ 
campestris pv
 \\ 
\textbf{\textcolor{blue}{Ogg1 protein}} \\ 
forward projection
 \\ 
preformed triplexes
 \\ 
\textbf{\textcolor{blue}{areA product}} \\ 
\textbf{\textcolor{blue}{alkaline phosphatase}} \\ 
Aryl hydrocarbon
 \\ 
CEN ENV \\
Epidemiologic Follow \\
\textbf{\textcolor{blue}{PKC beta}}
 \end{tabular} & 
 \begin{tabular}[c]{@{}l@{}}
\textbf{\textcolor{blue}{tyrosine kinase receptor}} \\ 
\textbf{\textcolor{blue}{tyrosine kinase}} \\ 
dystrophic epidermolysis
 \\ 
exacerbate cryoblobulinemia
 \\ 
fluoromethyl ketone
 \\ 
\textbf{\textcolor{blue}{uPA mRNA}} \\ 
police officers
 \\ 
\textbf{\textcolor{blue}{uPA mRNA}} \\ 
Enterococcus faecalis
 \\ 
Fugu rubripes
 \\ 
\textbf{\textcolor{blue}{ets family}} \\ 
\textbf{\textcolor{blue}{RNA polymerase}} \\ 
Nicotiana tabacum
 \\ 
P22 R17
 \\ 
San Francisco
 \\ 
\textbf{\textcolor{blue}{thymidine kinase promoter}} \\ 
bicycle ergometer
 \\ 
\textbf{\textcolor{blue}{paired domain}} \\ 
dura mater
 \\ 
fluticasone propionate
 \\ 
\textbf{\textcolor{blue}{recombinant human erythropoietin}} \\ 
\textbf{\textcolor{blue}{CAT reporter gene}} \\ 
orientational anisotropy
 \\ 
patent ductus
 \\ 
pia mater
 \\ 
\textbf{\textcolor{blue}{translation upstream factor}} \\ 
{\centering{
$\cdot$}}
 \\ 
{\centering{
$\cdot$}}
 \\ 
{\centering{
$\cdot$}}
 \\ 
epidermolysis bullosa
 \\ 
\textbf{\textcolor{blue}{fork head}} \\ 
\textbf{\textcolor{blue}{dopamine receptor}} \\ 
PCC 7120
 \\ 
Selected topics
 \\ 
chloromethyl alkyl
 \\ 
\textbf{\textcolor{blue}{firefly luciferase gene}} \\ 
irritation sensation
 \\ 
\textbf{\textcolor{blue}{viral LTR}} \\ 
Fusarium moniliforme
 \\ 
Jenkins Activity
 \\ 
\textbf{\textcolor{blue}{histone H3}} \\ 
Medical Radiology
 \\ 
\textbf{\textcolor{blue}{S1 nuclease}} \\ 
NnS neurones
 \\ 
Rhizobium leguminosarum
 \\ 
\textbf{\textcolor{blue}{9804 gene}} \\ 
\textbf{\textcolor{blue}{cyclin D1}} \\ 
emollient cream
 \\ 
imino protons
 \\ 
nontumorigenic Ad5
 \\ 
\textbf{\textcolor{blue}{MAP kinase}} \\ 
proportional hazards
 \\ 
\textbf{\textcolor{blue}{pertussis toxin}} \\ 
volatile solvents
 \\ 
Karger AG
 \\ 
\textbf{\textcolor{blue}{env genes}} \\ 
\textbf{\textcolor{blue}{integrin subunits}} \\ 
aryl hydrocarbon
 \\ 
dyad symmetry
 \\ 
multifocal leukoencephalopathy
 \end{tabular}  & 
 \begin{tabular}[c]{@{}l@{}}
\textbf{\textcolor{blue}{glucocorticoid receptor}} \\ 
ad lib
 \\ 
ad libitum
 \\ 
aggregative fimbriae
 \\ 
\textbf{\textcolor{blue}{thyroid hormone receptor}} \\ 
amylose cornstarch
 \\ 
\textbf{\textcolor{blue}{prolyl isomerase}} \\ 
fenfluramine anorexia
 \\ 
hexamethylpropyleneamine oxime
 \\ 
\textbf{\textcolor{blue}{IgG antibodies}} \\ 
\textbf{\textcolor{blue}{rheumatoid factor}} \\ 
myasthenia gravis
 \\ 
otoacoustic emissions
 \\ 
substantia innominata
 \\ 
\textbf{\textcolor{blue}{PDGF receptors}} \\ 
synovial chondromatosis
 \\ 
\textbf{\textcolor{blue}{LDL cholesterol}} \\ 
vena cava
 \\ 
Dirofilaria immitis
 \\ 
\textbf{\textcolor{blue}{alpha 2AP}} \\ 
\textbf{\textcolor{blue}{Cre recombinase}} \\ 
Spodoptera frugiperda
 \\ 
Zea mays
 \\ 
reticulocyte lysate
 \\ 
\textbf{\textcolor{blue}{polypyrimidine tract binding protein}} \\ 
BACTEC 9000
 \\ 
 {\centering{
$\cdot$}}
 \\ 
{\centering{
$\cdot$}}
 \\ 
{\centering{
$\cdot$}}
 \\ 
\textbf{\textcolor{blue}{TCR beta}} \\ 
\textbf{\textcolor{blue}{NMDA receptor}} \\ 
SELECTION CRITERIA
 \\ 
acoustic neuroma
 \\ 
acoustic startle
 \\ 
\textbf{\textcolor{blue}{exonuclease III}} \\ 
aphthous stomatitis
 \\ 
\textbf{\textcolor{blue}{SR family}} \\ 
flexor motoneurons
 \\ 
plan spared
 \\ 
\textbf{\textcolor{blue}{antithrombin III}} \\ 
\textbf{\textcolor{blue}{epidermal growth factor}} \\ 
rear corner
 \\ 
vas deferens
 \\ 
vinyl siloxane
 \\ 
\textbf{\textcolor{blue}{ERK MAPK}} \\ 
interspecific backcross
 \\ 
\textbf{\textcolor{blue}{growth hormone}} \\ 
SB 203580
 \\ 
circular dichroism
 \\ 
\textbf{\textcolor{blue}{beta receptor}} \\ 
\textbf{\textcolor{blue}{TK gene}} \\ 
hypoxaemic resuscitation
 \\ 
intraindividual fluctuations
 \\ 
northern Norway
 \\ 
\textbf{\textcolor{blue}{capsid proteins}} \\ 
prizidilol hydrochloride
 \\ 
\textbf{\textcolor{blue}{SH3 domain}} \\ 
thiazide diuretics
 \\ 
von Willebrand \\
\textbf{\textcolor{blue}{proliferating cell nuclear antigen}} \\ 
 \end{tabular} \\ \bottomrule
\end{tabular}
}
\caption{PMI vocabulary constructed from BC2GM dataset. DS-terms (in blue) discovered within the constructed PMI vocabularly}
\label{tab:pmi_vocab}
\end{table*}

\end{document}